\documentclass[sigconf]{acmart}
\setcopyright{none}
\renewcommand\footnotetextcopyrightpermission[1]{}
\usepackage{bbm}
\usepackage{algorithm}
\usepackage{algpseudocode}
\usepackage{graphicx}

\def\our{SplitEE}

\begin{document}

\title{SplitEE: Early Exit in Deep Neural Networks  with Split Computing
}

\author{Divya J. Bajpai, Vivek K. Trivedi, Sohan L. Yadav, and Manjesh K. Hanawal}
\affiliation{%
	\institution{MLiONS Lab, IEOR}
	\city{IIT Bombay, Mumbai}
	\country{India}}
\email{{mhanawal@iitb.ac.in, 203170005@iitb.ac.in}}

%
%

\begin{abstract}
Deep Neural Networks (DNNs) have drawn attention because of their outstanding performance on various tasks. However, deploying full-fledged DNNs in resource-constrained devices (edge, mobile, IoT) is difficult due to their large size. To overcome the issue, various approaches are considered, like offloading part of the computation to the cloud for final inference (split computing) or performing the inference at an intermediary layer without passing through all layers (early exits). In this work, we propose combining both approaches by using early exits in split computing. In our approach, we decide up to what depth of DNNs computation to perform on the device (splitting layer) and whether a sample can exit from this layer or need to be offloaded. The decisions are based on a weighted combination of accuracy, computational, and communication costs.
We develop an algorithm named \our{} to learn an optimal policy.
Since pre-trained DNNs are often deployed in new domains where the ground truths may be unavailable, and samples arrive in a streaming fashion, \our{} works in an online and unsupervised setup. 
We extensively perform experiments on five different datasets. 
\our{} achieves a significant cost reduction ($>50\%$) with a slight drop in accuracy ($<2\%$) as compared to the case when all samples are inferred at the final layer. The anonymized source code is available at \url{https://anonymous.4open.science/r/SplitEE_M-B989/README.md}.
\end{abstract}


\maketitle
\pagestyle{plain}

\section{Introduction}
Large-scale pre-trained Deep Neural Networks (DNNs) have gained attention due to their high accuracy over diverse tasks. However, increased accuracy comes at a cost of increased computational cost and latency due to their large number of parameters ( e.g.,  ELMO~\cite{peters1802deep} ($\sim$ 13.6M), BERT~\cite{devlin2018bert} ($\sim$ 110M), GPT~\cite{radford2019language} ($\sim$ 120M), XLNet~\cite{yang2019xlnet} ($\sim$ 110M) and RoBERTa~\cite{liu2019roberta}  ($\sim$ 123M) parameters.) Deploying such DNNs on resource-constraint mobile or other edge devices is challenging. Several methods are proposed to address the challenge of computational requirements of DNNs in edge devices, like Split computing, Early exits, and Offloading to the cloud \cite{matsubara2022split}.
\\
\noindent
{\bf Split computing:} Resource-constrained edge devices can offload data to the cloud, which typically has more processing capability (e.g., Graphical Processing Units). Full-fledged DNN can run on the cloud, and the inference can be returned to the edge devices. However, offloading samples from the edge to the cloud adds extra communication costs and latency. To better utilize the resources at edge and cloud, "split computing" is proposed, where a DNN model is split and implemented in two parts -- a few initial layers of DNN are deployed on the edge device, and the remaining layers are deployed on the cloud. This hybrid strategy allows for edge-cloud co-inference. In split computing, each sample passes through all layers of the DNN, and inference is completed at the final layer of the DNN on the cloud.  Though split computing reduces the computational load on the edge devices, the communication cost remains as in the cloud-only setup, as all the samples are offloaded to the cloud. 

 \noindent
{\bf Early exits:} Many DNNs, like BranchyNET \cite{teerapittayanon2016branchynet}, SPINN \cite{laskaridis2020spinn}, DeeBERT \cite{xin-etal-2020-deebert} and ElasticBERT \cite{liu2021elasticbert} allow inference at an intermediate layer to make an 'early-exit.'
In such early-exit DNNs, one has to decide at which layer to exit, and the decision is based on the tradeoff between accuracy and cost (computational, latency): exiting from early layers requires low computational resources but can suffer from imprecise predictions, while the sample exiting from later layers consume higher computational resource but can be predicted with improved accuracy. Deciding which layers to exit is critical in achieving the best tradeoff between accuracy and cost.

In this work, we propose combining split computing and early exit approaches to improve the resource utilization and cost incurred. In split computing, we refer to the last layer in the edge from which offloading to the cloud happens as the {\it splitting layer}. Before a sample is offloaded to the cloud, one can make an inference at the splitting layer to check. If the confidence in the inference is high, the offloading process can be aborted, and the sample can exit with the inference; otherwise, the normal offloading process can continue. The main challenge in our approach is in selecting the splitting layer, i.e., which should be the last layer of the DNN on the edge, and then deciding whether to offload or early exit from it so that accuracy and cost (computational and communication) are optimally balanced. 



Our proposal is also motivated by practical considerations; DNNs like BERT consist of multiple transformer layers, each being a replica of the other. From the hardware implementation point of view, the same hardware module is reused to realize each layer of the DNN -- after computation from a layer, its output parameters are fed back into its input to realize the computations of the next layer. Then one can realise the required number of DNN layers by repeating the computations in the hardware module many times. At the end of each iteration (or a layer), one can check confidence (entropy or probability of most likely class) in the prediction. If the confidence is high, the sample can exit without requiring further computation or being offloaded to the cloud. However, if the confidence improvement after passing through a few layers is not substantial, the sample can be offloaded to the cloud, where it can be processed till the last layer without resource constraints.



We develop a learning algorithm named \our{} - \underline{Split} Computing with \underline{E}arly \underline{E}xits.
We consider the setup where data arrives in an online fashion (sequential order), and the ground truth of the samples may not be available. Hence the algorithm has to work in the online unsupervised setup. As we do not assume the availability of any labels, we use confidence as a proxy for accuracy. SplitEE aims to maximize total confidence in predictions while considering computational and offloading costs. The computational costs capture the cost of running DNN layers on edge, and offloading cost captures the cost of communicating the DNN output from the splitting layer to the cloud. We define a reward function with a weighted difference between confidence and the cost incurred. We use the multi-armed bandit framework and set our objective as minimizing the expected cumulative regret, defined as the difference between the cumulative reward obtained by an oracle and that obtained by the algorithm. SplittEE is based on the classical Upper Confidence Bound (UCB) \cite{auer2002finite} algorithm. We also develop a variant of SplitEE that takes into account the additional information available at the intermediary levels in the form of side observations. We refer to this variant as SplitEE-S.  

We use the state-of-art Early-Exit DNN ElasticBERT for natural language inference as a test bed to evaluate the performance of our algorithm. ElasticBERT is based on the BERT backbone and trains multiple exits on a large text corpus. We extensively evaluate the performance of \our{} and SplitEE-S on five datasets \textit{viz.} IMDb, Yelp, SciTail, QQP and SNLI to cover three types of classification tasks -sentiment classification, entailment classification, and semantic equivalence classification. We first prepare an Early-Exit DNN by fine-tuning it on a similar kind of task to perform inference on an equivalent task with a different distribution in an unsupervised online manner. For instance, we fine-tune ElasticBERT on SST-2, a sentiment classification dataset and then evaluate \our{} on Yelp and IMDb which have review classification tasks.
\our{} finds an optimal splitting layer such that samples could be inferred locally only if the given sample meets the confidence threshold. In this way, \our{} only infers 'easy' samples locally forcing less load on mobile devices and offloads 'hard' samples. \our{} observes the smallest performance drop of $<2\%$ in the accuracy and $>50\%$ reduction in cost as compared to the case when all samples are inferred at the final exit. During inference, these DNNs might be applied to a dataset having different latent data distribution from the dataset used to train the DNN. The optimal splitting layer might be different depending on the latent data distribution. Hence \our{} adaptively learns the optimal split point by utilizing the confidence available at the exit attached to the splitting layer and computational cost.     

\begin{figure}
    \centering
    \includegraphics{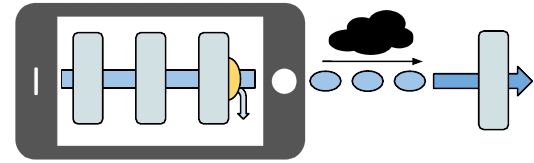}
    \caption{Efficient edge-cloud co-inference setup where part of the layers are executed on the edge device with an option to exit (infer a sample) at the split point and remain on the cloud to infer at the final layer.}
    \label{fig:eenn}
    \vspace{-.5cm}
\end{figure}

Our main contributions are as follows:
\begin{itemize}
    \item We introduce early exits in split computing and introduce a learning model. In our model, the decision is to find the split point as well as whether to exit or offload from the split point. 
    \item To find the optimal split point, we develop upper-confidence-based algorithms \our{} that decide the splitting layer on the fly without requiring any ground truth information and achieve sub-linear regret.

    \item We optimize the utilization of resources on edge and the cloud without sacrificing much accuracy by only inferring easy samples on edge devices.

    \item Using five distinct datasets, we empirically verify that our algorithms significantly reduce costs with a small drop in accuracy compared to the base-lines and state-of-the-art algorithms.
\end{itemize}

\section{Related works}
In this section, we discuss the previous works on Split Computing, Early-Exit DNNs and the utilization of DNNs on Mobile devices.

\subsection{Split Computing in DNNs}
Neurosurgeon \cite{kang2017neurosurgeon} searches for the best splitting layer in a DNN model by minimizing the cost associated with a splitting layer.
Split Computing is applied with different approaches. BottleNet \cite{eshratifar2019bottlenet} and Bottlefit \cite{matsubara2022bottlefit} introduce a bottleneck in split computing where part of the DNN in the mobile device will encode the sample into a reduced size. The reduced-sized sample is then offloaded. The sample is then decoded and inferred on the cloud. There are multiple training methodologies to encode the input on the mobile device. BottleNet++ \cite{shao2020bottlenet++} and \cite{hu2020fast} perform cross-entropy-based training, Matsubara \cite{matsubara2019distilled} perform knowledge-distillation based training, CDE \cite{sbai2021cut} and Yao \cite{yao2020deep} perform reconstruction-based training and Matsubara \cite{matsubara2021neural} perform head network distillation training method to effectively encode the input to offload efficiently.

\subsection{Early-exit Neural networks}
Early-exit DNNs are employed on various tasks. In image classification
BranchyNet \cite{teerapittayanon2016branchynet}, among other earlier research, uses classification entropy at each associated exit following each layer to determine whether to infer the sample at the side branch. If the exit point's entropy is below a predetermined threshold, the choice to exit is made. Similarly, SPINN \cite{laskaridis2020spinn} and SEE \cite{wang2019see} also use the estimated confidence measure at the exit branch to determine whether to exit early. However, the confidence estimate here is the likelihood of the most likely class. 

Besides exiting early, works like FlexDNN \cite{fang2020flexdnn} and Edgent \cite{li2019edge} focus mainly on the most appropriate DNN depth. Other works, such as Dynexit \cite{wang2019dynexit}, focus on deploying the multi-exit DNN in hardware. It trains and deploys the DNN on Field Programmable Gate Array (FPGA) hardware while Paul \textit{et al.} \cite{kim2020low} explains that implementing a multi-exit DNN on an FPGA board reduces inference time and energy consumption. 

In the NLP domain, DeeBERT \cite{xin2020deebert}, ElasticBERT \cite{liu2021elasticbert} and BERxiT \cite{xin2021berxit}  are transformer-based BERT models. DeeBERT is obtained by training the exit points attached before the last module to the BERT backbone separately while ElasticBERT trains the backbone with attached exits jointly with the final exit. BERxiT proposes a more advanced fine-tuning strategy for the BERT model with attached exits. PABEE\cite{zhou2020bert} and Pcee-BERT\cite{zhang2022pcee} suggest an early exit depending on the agreement between early-exit classifiers up to a fixed patience threshold. LeeBERT \cite{zhu2021leebert} on the other hand applies knowledge distillation across all exit layers rather than just distilling the knowledge prediction from the final layer.

\subsection{DNNs in Mobile Devices}
Pacheco \cite{pacheco2021early} utilize both multi-exit DNN and DNN partitioning to offload mobile devices via multi-exit DNNs. Similarly, EPNet \cite{dai2020epnet} learns when to exit considering the accuracy-overhead trade-off but in an offline fashion.

LEE \cite{ju2021learning}, DEE \cite{ju2021dynamic} and UEE-UCB \cite{hanawal2022unsupervised} utilize the multi-armed bandit framework to learn the optimal exit. However, they do not have the option to offload and infer only at mobile devices after finding the optimal exit. LEE and DEE provide efficient DNN inference tasks for mobile devices in scenarios such as service outages. Both LEE and DEE assume that the utility is revealed which depends on the ground truth labels.
LEE and DEE use the classical UCB1 \cite{auer2002finite} algorithm to learn the optimal exit. UEE-UCB learns the optimal exit in a setup similar to ours, however, it does not have the option to offload. It finds the optimal splitting layer and infers all the samples through the mobile device. It also assumes that the intermediary layers follow the strong dominance property.

Following are major differences between our setup in comparison with the previous setups: 1) We take into account both the computational and communication costs in addition to accuracy in deciding the splitting layer, whereas the previous works on split computing considered only the communication cost, while the early exit work considered only computational costs along with the accuracy. 2) Our work is completely in an unsupervised online setup as it does not require any ground truth information.  3) For each sample, we use the contextual information (confidence) to decide whether to exit or offload at the splitting layer dynamically.  Table \ref{tab: Results} provides a direct comparison to state-of-arts. 

\section{Problem setup}\label{sec:problem setup}
We are given a pre-trained DNN with $L$ layers with attached exits after every layer. We index the layers using the set $[L]  =\{1,2,\ldots L\}$. We consider classification tasks with a target class $\mathcal{C}$.
For an input $x$ and layer $i\in [L]$, let $\hat{P}_i(c)$ denote the estimated probability that $x$ belongs to class $c\in \mathcal{C}$. Let $C_i =  \max_{c\in \mathcal{C}}\hat{P}_i(c)$ denote the confidence of estimated probability class. Input $x$ is processed sequentially through the DNN. The DNN could be split at any layer $i \in [L]$, where the layers $1,2,\ldots,i$ are on the mobile device and the remaining layers, i.e., $i+1, i+2, \ldots, L$ are on the cloud.  In our setup for each sample following two-stage decisions has to be made
1) Where to split the DNN?
2) Whether to exit from the splitting layer offload to the cloud. The decision on where to split the DNN does not depend on the individual samples but on the underlying distribution. Whereas the decision to offload or exit is based on each sample as follows: 
If the split is at the $i$th layer, $C_i(x)$ is computed and compared against a pre-defined threshold $\alpha$. If $C_i(x)\geq \alpha$, the sample exits and is inferred on the mobile device at the splitting layer, otherwise it is offloaded and inferred at the final layer on the cloud. 

The cost of using the DNN up to layer $i$ could be interpreted as the computational cost of processing the sample till layer $i$ and performing inference. Let $\gamma_i$ be the cost associated with the split performed at the $i$th layer. We set  $\gamma_i \propto i$ as the amount of computation that depends on the depth of the splitting layer in the DNN. We denote the cost of offloading from mobile to cloud as $o$. The value of $o$ across all layers depends on the size of the input and the transmission cost (\textit{e.g.} Wi-Fi, 5G, 4G and 3G). We define the reward when the splitting is performed at the $i \in [L]$ layer as
\begin{equation}
\label{eq: Reward1}
    r(i) = \left\{
        \begin{array}{ll}
            C_i-\mu\gamma_i & \textit{if} \quad C_{i} \geq \alpha \text{ or } i=L\\
             C_L- \mu(\gamma_i+ o) & \textit{otherwise,}
        \end{array}
    \right.
\end{equation}
where $\mu$ is a conversion factor to express the cost in terms of confidence. $\mu$ is input by the users depending on their preference for accuracy and computational costs.

The reward could be interpreted as follows: if the DNN is confident on the sample in the prediction obtained from the $i$th layer, then the reward will be the gain in confidence subtracted by the cost of moving the sample till $i$th layer and inferring. If not, then the sample is offloaded to the cloud for inference, where the confidence of $C_L$ is achieved at the last layer, and an additional offloading cost $o$ is incurred. Observe that if $i = L$, all the computations are executed on the edge device, and the sample is inferred at $L$th layer (without offloading). We define $i^{*} = \arg\max_{i\in[L]}\mathbb{E}[r(i)]$ which is defined as for $i\in[L-1]$ as
\begin{multline}
\mathbb{E}[r(i)] = \mathbb{E}[C_i-\mu\gamma_i|C_i\geq \alpha]\cdot P[C_i\geq\alpha]\\ + \mathbb{E}[C_L-\mu(\gamma_i+o)|C_i<\alpha]\cdot P[C_i<\alpha],
\end{multline}
and for the last layer $L$, it is a constant given as $\mathbb{E}(r(L)) = C_L-\mu\gamma_L$.
The goal is to find an optimal splitting layer $i^{*}$ such that sample will be inferred at $i^{*}$ or be offloaded to the cloud for inference.

We model the problem of finding the optimal splitting layer as a multi-armed bandit problem (MAB). We define the action set as layer indices in the DNN $\mathcal{A} = [L]$. Following the terminology of MABs, we also refer to elements of $\mathcal{A}$ as arms. Consider a policy $\pi$ that selects arm $i_t$ at time $t$ based on past observations. We define the cumulative regret of $\pi$ over $T$ rounds as 
\begin{equation}
    R(\pi, T) = \sum_{t=1}^{T}\mathbb{E}[r(i^{*})-r(i_t)]
\end{equation}
where the expectation is with respect to the randomness in the arm selection caused by previous samples. A policy $\pi^{*}$ is said to be sub-linear if average cumulative regret vanishes, i.e., $R(\pi^{*}, T)/T\rightarrow 0$. We experimentally prove that both variants of Algorithm \ref{alg:algorithm} achieves sub-linear regret.

\section{Algorithm}
\begin{figure*}
    \centering

ops    \includegraphics[height=5cm, width=16cm]{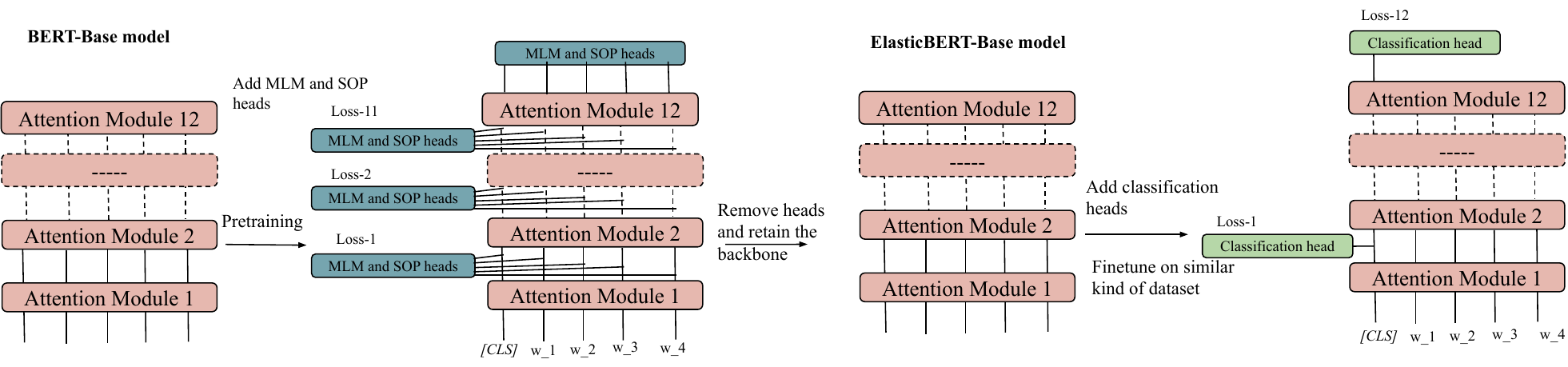}
    \caption{Pre-training a multi-exit DNN}
    \label{fig:ElasticBERT}
\end{figure*}
In this section, we develop an algorithm named \underline{Split} computing with \underline{E}arly \underline{E}xit (SplitEE). The algorithm is based on the `optimism in the face of the uncertainty principle' and uses the upper confidence bounds.  In this variant, the inference is performed only at the splitting layer, and the decision to offload or exit is based on confidence in this inference. In the following subsection, we develop another variant named SplitEE-S that makes inferences at each layer and not just at the splitting layer.  

\subsection{\our{}}
The input to this variant is the confidence threshold ($\alpha$), the exploration parameter ($\beta$), the number of layers ($L$), and the computational cost for each layer $\gamma$ which could be split as $\gamma = \lambda_{1}+\lambda_{2}$ where $\lambda_{1}$ could be interpreted as the processing cost whereas $\lambda_2$ is the inference cost at the attached exit. Since we are not utilizing the exits attached to the layer before the chosen splitting layer, hence in this variant, $\lambda_2$ will only be accumulated for the splitting layer selected.
\begin{algorithm}[H]
        \caption{\our{}}
        \begin{algorithmic}[1]
          \State\textbf{Input:} $\alpha \text{ (threshold)}, \beta\geq 1, L,\text{ cost } \gamma_i \text{ } \forall i\in[L]$
\State \textbf{Initialize:} $Q(i)\gets 0, N(i)\gets 0$.
\State Initialize by playing each arm once.
\For{$t = |L|+1, |L|+2, \ldots$}
\State Observe an instance $x_t$ 
\State $i_t \gets \arg \max_{i\in \mathcal[L]}\left(Q(i)+\beta\sqrt{\frac{\ln(t)}{N(i)}}\right)$
\State Pass $x_t$ till layer $i_t$, apply threshold $\alpha$ and observe $C_{i_t}$

\If{$C_{i_t}\geq \alpha$} 
    \State  Infer at layer $i_t$ and exit
    \State $r_t(i_t) \gets C_{i_t}(x_t)-\mu\gamma_{i_t}$, $N_{t}(i_t) \leftarrow N_{t-1}(i_t)+1$
    \State $Q_{t}(i_t) \leftarrow \sum_{j=1}^{t}r_{j}(k)\mathbbm{1}_{\{k=i_t\}}/N_{t}(i_t)$ 
\Else 
    \State Offload to the last layer. Observe $C_L$
    \State $r_{t}(i_t) \gets C_{L}(x_t)-\mu(\gamma_{i_t} + o)$, $N_{t}(i_t) \leftarrow N_{t-1}(i_t)+1$
    \State $Q_{t}(i_t) \leftarrow \sum_{j=1}^{t}r_{j}(k)\mathbbm{1}_{\{k=i_t\}}/N_{t}(i_t)$  
    \State Infer at the last layer
\EndIf
\EndFor
\end{algorithmic}
\label{alg:algorithm}
\end{algorithm}
The pseudo-code of the SplitEE is given in Algorithm \ref{alg:algorithm}. The algorithm works as follows:
It plays each arm once for the first $L$ instances to obtain the rewards $Q(i)$ and the counters $N(i)$ for each layer once. 
Then it plays an arm $i_t$ that maximises the UCB index (line 6) in succeeding rounds. The weighted sum of the empirical average of the rewards $Q(i)$ and confidence bonuses is used to create UCB indices. If the confidence at layer $i_t$ is above the threshold $\alpha$, the sample exits the DNN; else it is offloaded to the cloud with additional cost $o$. Following analysis of UCB1\cite{ML02_UCB1_Auer}, one can verify that SplitEE achieves a sub-linear regret. More specifically the regret is $\mathcal{O}\left(\sum_{\alpha \in \mathcal{A}_p\backslash\alpha^{*}}\frac{log(n)}{\Delta_{\alpha}}\right)$ where $\Delta_i = r(i^*)-r(i)$ denotes the optimality gap.

\subsection{\our{}-S}
As the sample passes through each layer, inference can be performed and confidence in the prediction can be evaluated. Thus when the sample reaches the splitting layer, confidence in predictions of all the previous layers will be available. This can be used as side observations of other arms in the algorithm to accelerate the learning process. However, note that the side observation comes at the cost of performing inference $\lambda_2$ at each intermediary layer the sample passes through in the edge device, which will add up as an additional cost. We refer to this variant of SplitEE as SplitEE-S  (Split computing with early exit and side information). In this variant, the accumulated inference cost $\lambda_2$ will be higher than SplitEE as the inference is performed after every layer in the edge device. The computational cost will be the same in both variants.

SplitEE-S is the same as SplitEE except that the computations in lines 8-16 of Algorithm \ref{alg:algorithm} is performed for all $i\leq i_t$. Hence, for a sample, SplitEE-S updates the rewards for $|[i_t]|$ arms. Due to multiple updates, the convergence of the algorithm to the optimal arm becomes fast and the cumulative regret decreases. We experimentally verify the claim that the cumulative regret of SplitEE-S is smaller than that of SplitEE in fig. \ref{fig:Regret_curve_1}.


\section{Experiments}

\begin{figure*}
    \centering \includegraphics[scale = 0.279]{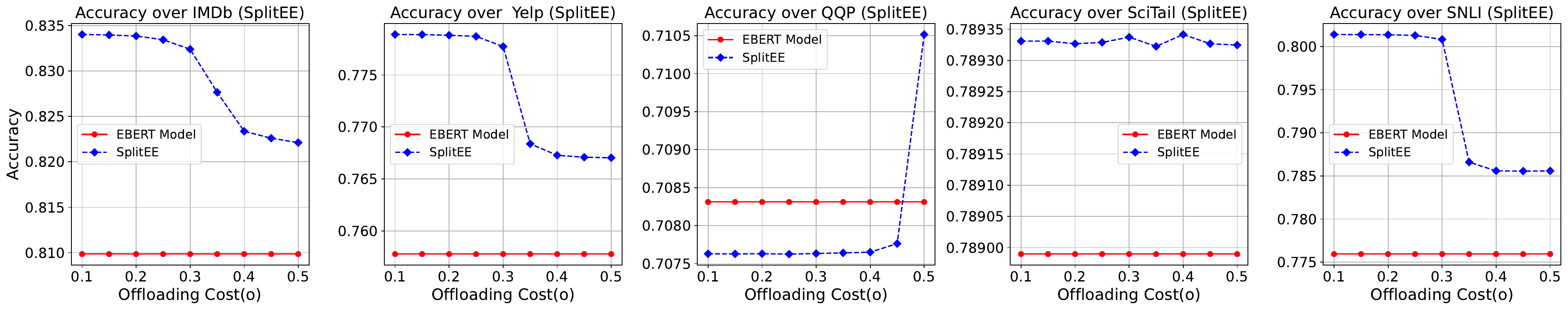}
    \caption{Accuracy for different offloading costs ($o$) (\our{})}
    \label{fig:Accuracy_res_eesplit}
\end{figure*}

\begin{figure*}
    \centering \includegraphics[scale = 0.279]{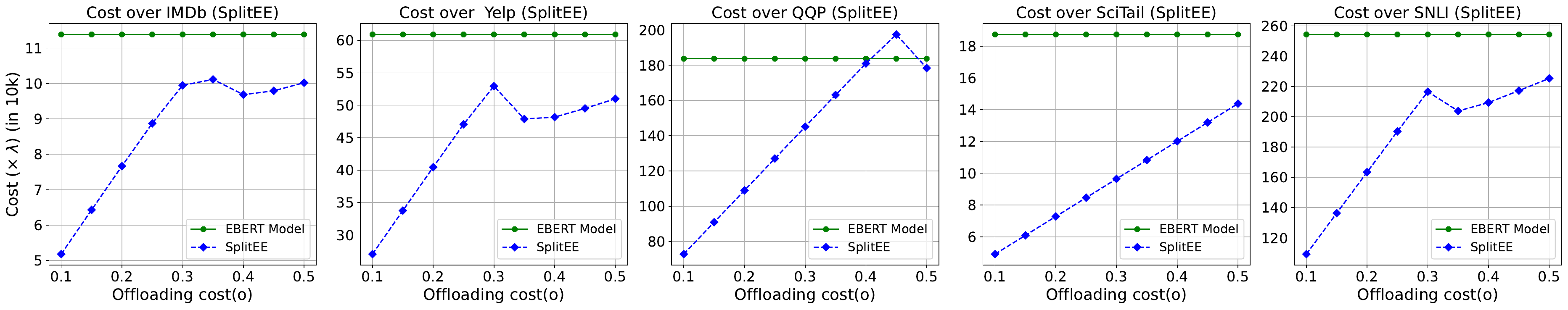}
    \caption{Cost (in $10^4\times\lambda$ units) for different offloading cost (\our{})}
    \label{fig:cost_res_eesplit}
\end{figure*}
In this section, we first explain how to prepare an Early-Exit DNN and then provide the experimental details.
\subsection{Preparing a Multi-Exit DNN}\label{sec: ElasticBERT}
To evaluate \our{}, we require an early-exit neural network that has already been trained. We use the ElasticBERT-base backbone, which is based on the 12-transformer-layer BERT-base model. For anytime inference, ElasticBert has exits attached after each transformer layer. After attaching exits, the model is trained on a large text corpus with joint masked language modelling (MLM) and sentence order prediction (SOP) loss functions across all exits. After discarding all exits, the ElasticBERT-base model's backbone with learnt weights remains. For more details of the training procedure, we refer to \cite{liu2021elasticbert}.

After having a pre-trained model backbone, we attach task-specific exits (\textit{e.g.} classification heads) after all transformer layers along the backbone and fine-tune it using a labeled dataset (from a similar domain to the evaluation dataset). Sentence-level representations for sentence-level tasks are learned using the [\textit{CLS}] token. After each transformer layer, this token representation is connected to the classification heads. 
A sketch of the entire training procedure is provided in figure \ref{fig:ElasticBERT}. $w_1, w_2, w_3$ and $w_4 $ represent token embeddings of the given input sequence. The head is attached to produce a representation that can be compared with the task label to compute the loss. Cross-entropy loss is the loss function that we select. Using learnable weights, the attached classification heads transform the [\textit{CLS}] token's \textit{q}-dimensional vector representation into a probability vector for direct comparison with the task label.

\begin{table}
\caption{Information about the size of datasets. FT data is the dataset used to prepare the ElasticBERT backbone for the corresponding task and \#Samples is the number of samples in the dataset. E.data is the evaluation dataset.}
\begin{tabular}{|l|l|l|l|}
\hline
\textbf{E. Data} & \textbf{\#Samples} & \textbf{FT Data} & \textbf{\#Samples} \\ \hline
IMDb             & 25K                & SST-2            & 68K                \\ \hline
Yelp             & 560K               & SST-2            & 68K                \\ \hline
SciTail          & 24K                & RTE              & 2.5K               \\ \hline
QQP              & 365K               & MRPC             & 4K                 \\ \hline
SNLI             & 550K               & MNLI             & 433K               \\ \hline
\end{tabular}

\label{tab: dataset} 
\end{table}

\subsection{Experimental setup}\label{sec: exp_setup}
In this section, we explain the experimental setup and details of \our{}.
We have three major steps in the experimental setup which are summarized below:

\textbf{i) Training the backbone:} Initially, we train the ElasticBERT-base model with MLM and SOP heads attached after every transformer layer of the BERT-base model. After training on a large text corpus, we remove the MLM and SOP heads from the ElasticBERT model, leaving only the backbone. We directly import weights of the learned backbone, hence this part does not need any computation.

\textbf{ii) Fine-tuning and learning weights (Supervised):} In the backbone obtained by step (i), we attach task-specific exits (heads) after each transformer layer, and to learn weights for these heads, we perform supervised training. Since we assume that we do not have labels for the evaluation task. We utilize a labeled dataset with a similar kind of task but with a different distribution or from a different domain. For example, we evaluate \our{} on IMDb and Yelp datasets which are review classification datasets and learn the weights for heads using the SST-2 dataset which has a similar task of sentiment classification but with different latent data distribution. 

\textbf{iii) Online learning of optimal splitting layer (Unsupervised):} Finally we use weights from step (2) to learn the optimal splitting layer in an unsupervised and online setup for the evaluation tasks. We perform this step after the model has been deployed and ready for inference.

\begin{figure*}
    \centering \includegraphics[scale = 0.279]{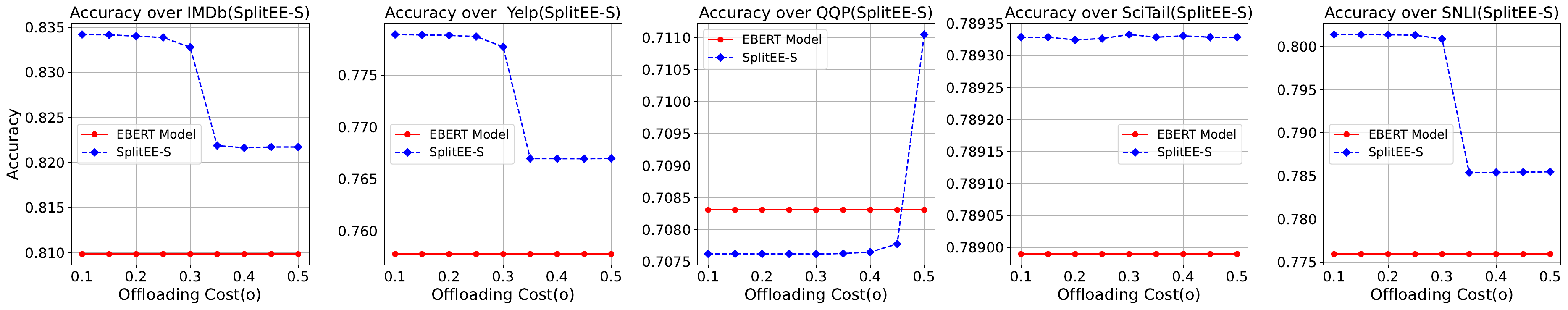}
    \caption{Accuracy for different offloading costs ($o$) (\our{}-S)}
    \label{fig:Accuracy_res_EESPlit s}
\end{figure*}

\begin{figure*}
    \centering \includegraphics[scale = 0.279]{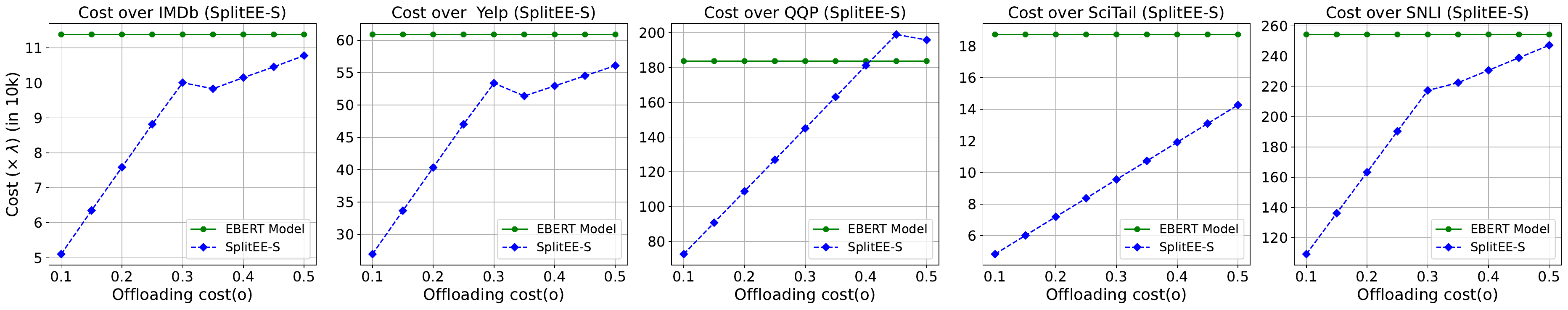}
    \caption{Cost (in $10^4\times\lambda$ units) for different offloading cost (\our{}-S)}
    \label{fig:cost_res_EESPLIT-S}
\end{figure*} 

We perform experiments on single NVIDIA RTX 2070 GPU. Part (i) does not require any computation as we can directly import the weights from the backbone. Part (ii) takes a maximum of 10 GPU hour of runtime (on the MNLI dataset). Part (iii) is not computationally involved and could be executed in $<1$ hour of CPU runtime and does not requires GPU support on a single run.

We evaluated \our{} on five datasets covering three types of classification tasks. The datasets used for evaluation are:
\begin{enumerate}
    \item \textbf{Review classification on IMDb \cite{maas2011learning} and Yelp \cite{asghar2016yelp}:} IMDb is a movie review classification dataset and Yelp consists of reviews from various domains such as hotels, restaurants etc. For these two datasets, ElasticBERT is finetuned on \textbf{SST-2 (Stanford Sentiment classification)} dataset which is also a sentiment classification dataset.
    \item \textbf{Entailment classification on SciTail:} SciTail is an entailment classification dataset created from multiple questions from science and exams and web sentences. To evaluate SplitEE on SciTail, it is finetuned on \textbf{RTE(Recognizing Textual Entailment)} dataset which is also an entailment classification dataset.
    \item \textbf{Entailment classification on SNLI(Stanford Natural Language Inference) (Multi-class):} SNLI is a collection of human-written English sentence pairs manually labelled for balanced classification with labels \textit{entailment}, \textit{contradiction} and \textit{neutral}. For evaluation of this dataset, ElasticBERT is finetuned on \textbf{MNLI(Multi-Genre Natural Language Inference)} which also contains sentence pairs as premise and hypothesis, the task is the same as for SNLI.

    \item \textbf{Semantic equivalence classification on QQP(Quora Question Pairs)}: QQP is a semantic equivalence classification dataset which contains question pairs from the community question-answering website Quora. For this task, we finetuned ElasticBERT on \textbf{MRPC(Microsoft Research Paraphrase Corpus)} dataset which also has a semantic equivalence task of a sentence pair extracted from online news sources.
\end{enumerate}
  Details about the size of these datasets are in table \ref{tab: dataset}. Observe from the table that the size of the dataset used for fine-tuning is much smaller as compared to the size of the corresponding evaluation dataset. We do not split the evaluation dataset. Except for IMDb and Yelp, other datasets are a part of ELUE \cite{liu2021elasticbert} and GLUE \cite{wang2019glue} benchmark datasets.

We attach exits after every transformer layer in the ElasticBERT model. The predefined threshold $\alpha$ is directly taken from the ElasticBERT model which utilizes the validation split of fine-tuning data to learn the best threshold. The choice of action set depends on the number of layers of the DNN being used. The action set is $\mathcal{A} = [L]$ and for ElasticBERT $L = 12$. 
We have two types of costs: Computational cost and Offloading cost. As explained in section \ref{sec:problem setup}, the computational cost is proportional to the number of layers processed i.e. $\gamma_i = \lambda i$ where $\lambda$ could be interpreted as per-layer computational cost. We can split $\lambda = \lambda_1+\lambda_2$, where $\lambda_1$ and $\lambda_2$ resemble the per-layer processing cost and per-layer inference cost respectively. We relate $\lambda_1$ and $ \lambda_2$ in terms of the number of matrix multiplications required to process and infer. We observe that $\lambda_2 = \lambda_1/6$ (5 matrix multiplications are needed for processing and 1 for inferring). Hence the cost for SplitEE-S will be $\lambda i $ and $\lambda_1 i + \lambda_2$ for SplitEE if $i$th layer is chosen as the splitting layer.
Since offloading cost is also user-defined and depends on the communication network used (e.g. 3G, 4G, 5G and Wi-Fi). Hence in the experiments, we provide results on different offloading costs $o$ from the set $\{\lambda, 2\lambda, 3\lambda, 4\lambda, 5\lambda\}$ as it is user-defined. With increasing stages of broadband mobile communication powers, we observe that offloading cost is at most five times the per-layer computational cost. For more details on how to compute the offloading cost, we refer to \cite{kuang2019partial}. For table \ref{tab: Results}, we use the fixed offloading cost as $o = 5\lambda$ (highest offloading cost). Without loss of generality, we choose $\lambda = 1$ for conducting all the experiments. The cost accumulated is however left in terms of $\lambda$ which is user-specific as all the cost is now in terms of $\lambda$.


The trade-off factor $\mu$ ensures that the reward function gives similar preferences to cost as well as accuracy. For the algorithm, we choose $\mu = 0.1$ to directly compare confidence and cost.
We repeat each experiment $ 20$ times and in each run the samples are randomly reshuffled and then fed to the algorithm in an online manner. In each round, the algorithm chooses a splitting layer and accumulates the regret if the choice is not optimal. We plot the expected cumulative regret in figure \ref{fig:Regret_curve_1}. The accuracy and cost reported for \our{} and \our{}-S is computed considering the chosen splitting layer prediction in each round (i.e. for every sample) and then per-sample averaged for $20$ runs.

\begin{table*}[]
\caption{Main Results: Results on different baselines across different datasets. Cost is left in terms of $10^{4}\times \lambda$ units. $\lambda$ is user-defined. The offloading cost is taken 5$\lambda$ (worst-case).}
\label{tab: Results}
\begin{tabular}{ccccccccccc}
\hline
Model/Data  & \multicolumn{2}{c}{IMDb}         & \multicolumn{2}{c}{Yelp} & \multicolumn{2}{c}{SciTail} & \multicolumn{2}{c}{SNLI} & \multicolumn{2}{c}{QQP} \\ \hline
            & Acc           & Cost             & Acc            & Cost    & Acc             & Cost      & Acc            & Cost    & Acc           & Cost    \\ \hline
Final-exit  & 83.4          & 30.0             & 77.8           & 161.0   & 78.9            & 28.3      & 80.2           & 659.2   & 71.0          & 436.6   \\
Random-exit & -1.4          & -31.3\%          & -1.2  & -38.0\% & -0.7            & -31.8\%   & -2.0           & -41.5\% & -0.1          & -14.8\% \\
DeeBERT     & -6.1          & -43.3\%          & -2.5           & -59.0\% & -3.6            & -5.3\%    & -3.5           & -38.9\% & -6.7          & -50.1\% \\
ElasticBERT & -2.5          & -62.3\%          & -2.1           & -62.1\% & -0.1            & -40.2\%   & -2.7           & -61.4\% & -0.2          & -57.9\% \\
\our{} & -1.3 & \textbf{-66.6}\% & -1.1 & \textbf{-68.3}\% & 0.0 & -49.2\% & -1.6 & \textbf{-65.8}\% & -0.1 & \textbf{-59.1}\%\\
\our{}-S  & \textbf{-1.2} & -64.3\% & \textbf{-1.1}           & -65.2\% & \textbf{0.0}    & \textbf{-50.5\%}   & \textbf{-1.7}  & -62.5\% & \textbf{+0.1}  & -55.1\% \\ \hline
\end{tabular}
\end{table*}

\subsection{Baselines}
    \textbf{1) DeeBERT:} Similar to our setup, we fine-tune DeeBERT and then perform inference on the evaluation dataset.
    DeeBERT prepares the early exit model in two steps: (1) It learns the general weights and embeddings for the BERT backbone using the loss function attached only at the final layer, this part is similar to BERT fine-tuning. (2) After freezing the weights, it attaches a loss function after every transformer layer except the final layer. Note that DeeBERT does not have the option to offload. DeeBERT uses the entropy of the predicted vector as confidence. We fine-tune the entropy threshold in a similar fashion as used by DeeBERT. Since it does not make any difference, hence we keep the confidence of DeeBERT as the entropy of the predicted vector. Other parameters are kept the same as used by DeeBERT.

\textbf{2) ElasticBERT:} is also based on the BERT-base model, the only difference is ElasticBERT is jointly trained by attaching MLM and SOP heads after every transformer layer, Once the model is trained, it removes the heads leaving the backbone. More details are in section \ref{sec: ElasticBERT} and figure \ref{fig:ElasticBERT}. All the parameters are kept the same as the ElasticBERT setup.

\textbf{3) Random selection:} In random selection, we select a random exit point and then process the sample till chosen exit, if the confidence at chosen exit is above the threshold then exit, else offload. Then we calculate the cost and accuracy. We report the average accuracy and cost by running the above procedure 20 times.

\textbf{4) Final exit:} In this case, we process the sample till the final layer for inference. This setup has a constant cost of $\lambda L$. This baseline is similar to the basic inference of neural networks. We also utilize this setup for benchmarking.

\vspace{-0.5cm}
\subsection{Need for offloading}
As explained in section \ref{sec: exp_setup}, the maximum possible offloading cost is $5$-times the per-layer computational cost. Hence if a sample is not gaining sufficient confidence for classification till a pre-specified layer, we might want to offload it to the cloud. Previous methods process the sample throughout the DNN until it gains sufficient confidence. We observed that processing a sample beyond $6$th layer was accumulating more processing cost than the offloading cost. While experimenting, we marked that on average DeeBERT processes $51\%$ samples and ElasticBERT processes $35\%$ samples beyond $6$th exit layer. These many samples accumulate a large computational cost for edge devices. Since edge devices have fewer resources available, both DeeBERT and ElasticBERT might exploit these resources in terms of battery lifetime depletion and device lifetime.

While our setup decides on a splitting layer as a sample arrives. The sample is processed till chosen layer and if the sample gains sufficient confidence it exits the DNN else it offloads to the cloud for inference reducing cost drastically. Additionally, offloading helps in increasing accuracy as the last layer provides more accurate results on samples that were not gaining confidence initially.

\begin{figure*}
    \centering
    \includegraphics[scale = 0.279]{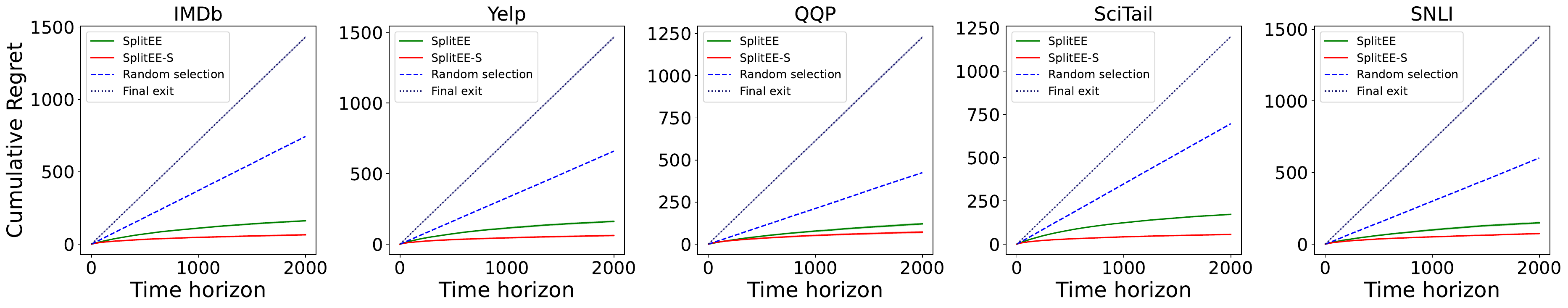}
    \caption{Regret for different models}
    \label{fig:Regret_curve_1}
\end{figure*}

\subsection{\our{} \ and \our{}-S}
From figures 
\ref{fig:Accuracy_res_eesplit},
\ref{fig:cost_res_eesplit}, \ref{fig:Accuracy_res_EESPlit s} and \ref{fig:cost_res_EESPLIT-S}, we observe that \our{} and \our{}-S have comparable performances. However, observe that \our{} does not utilize confidences of exits prior to chosen splitting layer. Hence we can directly process the sample to the splitting layer reducing the inference cost after each exit. \our{}-S uses the confidence available at all exits on the edge device to update rewards for multiple arms. The difference between the two is more evident in the regret curve (see fig.\ref{fig:Regret_curve_1}). \our-S curve saturates much earlier than \our{}. We also observed from the experiments that \our{} has a larger impact on the reshuffling of the dataset while \our{}-S is more robust to the reshuffling of the dataset as it needs less number of samples to learn the optimal splitting layer. Since in a real-world scenario, the size of the evaluation dataset might be small and we might need to adapt to changes in the distribution of data fast, in this case, SplitEE-S can be used. However, if the major concern is cost then \our{} will work better as it further reduces the inference cost (see table \ref{tab: Results}). Note that DeeBERT and ElasticBERT also incur the inference cost at all exits up to which a sample is processed.

\subsection{Analysis with different offloading cost}\label{sec:offloading}
Being user-defined, we analyse the behaviour of accuracy and cost on different offloading costs. Except for the QQP dataset, we observe a drop in accuracy as we increase the offloading cost. We explain the drop as more offloading cost will force more samples to exit early by choosing a deeper exit in DNN. More samples exiting early will make less accurate predictions.
Hence initially, when offloading cost is small it exits most samples from initial exits and offloads very few samples. When we increase offloading cost again it just goes deeper to gain confidence for those samples which were offloaded earlier. In terms of cost, it is evident that the cost of \our{} will go up as we increase the offloading cost. For the QQP dataset, we observe a reverse behaviour as there are very less samples that offload for QQP. We observed that there were many samples that exited the initial layers with miss-classifications (which is also an explanation for the lower cost of ElasticBERT).  As we increase the offloading cost \our{} looks for deeper exit layers to split hence a gain in accuracy.
Still, we are always better in terms of cost as well as accuracy when compared to ElasticBERT as shown in figure \ref{fig:cost_res_eesplit}, \ref{fig:cost_res_EESPLIT-S}, \ref{fig:Accuracy_res_eesplit} and \ref{fig:Accuracy_res_EESPlit s}. Detailed results are in table \ref{tab: Results}.


\subsection{Regret Performance}\label{sec: regret}
We repeat each experiment $20$-times. Each time, a randomly reshuffled data is fed in an online manner to the algorithm. 
In each step, the algorithm selects a splitting layer and accumulates the regret if the choice is not optimal. In figure \ref{fig:Regret_curve_1}, we plot the expected cumulative regret along with a 95\% confidence interval. We choose the exploration parameter $\beta = 1$. While each plot shows the results for a specific dataset. \our{} and \our{}-S outperforms the considered alternatives, yielding a lower cumulative regret and achieving sub-linear regret growth. We also observe that the \our{}-S achieves lower regret than the \our{}. This is because the side information provides the algorithm with additional information about the environment, which can be used to learn the optimal splitting layer quickly. As a result, the algorithm with side information can converge to the optimal policy more quickly. As observed from the figure \ref{fig:Regret_curve_1}, the regret starts saturating after the first 2000 samples for \our{} and after 1000 samples for \our{}-S.

\section{Results}

In table \ref{tab: Results}, we report the accuracy and cost across different datasets as well as different models. \our{} achieves smallest performance drop with a performance drop of $(<2\%)$ against the final-exit and largest depreciation in cost $(>50\%)$ as compared to final-exit. For the SciTail dataset, we are getting same accuracy as the final layer. This behaviour is observed as for most of the samples in SciTail, the gain in confidence is not sufficient in the initial layers, hence \our{} offloads most of the samples and achieves similar performance. It achieves a smaller cost than other baselines since DeeBERT and ElasticBERT process every sample to deeper exits to meet the confidence threshold and accumulate more cost. We observed that in QQP $15-20\%$ samples were misclassified with high confidence. Hence ElasticBERT exits many samples at initial layers but with a miss-classification incurring a lower cost.  However, we have gained accuracy from the final layer. The lower accuracy at final layer is the effect of overthinking\footnote{overthinking in inference is similar to over-fitting in training.} during inference. in general, the higher costs of DeeBERT and ElasticBERT could be explained as they process the sample till deeper exits until the sample's confidence is above a given threshold. However, \our{} suggests offloading if the sample does not gains sufficient confidence till the splitting layer. Accuracy of \our{} is also consistently higher as we also utilize the final layer for inference in conjunction with the splitting layer. Since the accuracy of the final exit is better than that of intermediate ones, \our{} achieves higher accuracy than other baselines.

\section{Conclusion}
We addressed the problem of using DNNs in resource-constraint edge devices like Mobile and IoTs. We proposed a new using mobile-cloud co-inference by combining Split computing and Early exits both of which are independently proposed to address the problem of deploying DNNs in resource-constrained environment. In our approach, part of DNN is deployed on the resource-constraint edge device and the remaining on the cloud. In the last layer of DNN implemented on the edge device, we make the inference, and depending on confidence in the inference, the sample either makes an exit or offloads to the cloud. The main challenge in our work is to decide where to split the DNN so that it achieves good accuracy while keeping computational and communication costs low. We developed a learning algorithm named SplitEE to address these challenges using the multi-armed bandit framework by defining a reward that takes into account accuracy and costs. Also, in our setup ground truth labels are not available. Hence SplitEE works in an unsupervised setting using confidence in prediction as a proxy for accuracy. Our experiments demonstrated that SplitEE achieves a significant reduction of cost (up to ~ 50 \%) with a slight reduction in accuracy (less than ~ 2 \%). We also developed a variant of SplitEE that exploits the side observation to improve performance.  

Our work can be extended in several ways. One \our{} assumed that the threshold used to decide whether to exit or offload is fixed based on offline validation. However, this can be adapted based on the new samples and can be made a learnable parameter. Also, in our work, we looked at an optimal split across all the samples. However, that can also be adaptive based on the sample.  Each sample is of a different difficulty level and deciding split based on its difficulty can further improve the prediction accuracy while still keeping the cost low. 

\section*{Acknowledgments}
Manjesh K.Hanawal thanks funding support from SERB, Govt of India, through Core Research Grant (CRG/2022/008807) and MATRICS grant (MTR/2021/000645). 

\bibliographystyle{ACM-Reference-Format}
\bibliography{sample-base}

\end{document}